\pdfoutput=1

\documentclass[11pt]{article}

\usepackage[final]{acl}

\usepackage{times}
\usepackage{latexsym}

\usepackage[T1]{fontenc}

\usepackage[utf8]{inputenc}
\usepackage{tipa}

\usepackage{microtype}

\usepackage{inconsolata}

\usepackage{enumitem}

\usepackage{csquotes} 
\usepackage{booktabs} 
\usepackage{tablefootnote} 

\usepackage{graphicx}

\usepackage{listings}
\definecolor{background}{HTML}{EEEEEE}
\definecolor{delim}{RGB}{20,105,176}
\definecolor{numb}{RGB}{106, 109, 32}
\definecolor{string}{rgb}{0.64,0.08,0.08}

\lstdefinelanguage{json}{
    numbers=none, 
    numberstyle=\small,
    frame=single,
    rulecolor=\color{black},
    showspaces=false,
    showtabs=false,
    breaklines=true,
    postbreak=\raisebox{0ex}[0ex][0ex]{\ensuremath{\color{gray}\hookrightarrow\space}},
    breakatwhitespace=true,
    basicstyle=\ttfamily\small,
    upquote=true,
    morestring=[b]",
    stringstyle=\color{string},
    literate=
     *{0}{{{\color{numb}0}}}{1}
      {1}{{{\color{numb}1}}}{1}
      {2}{{{\color{numb}2}}}{1}
      {3}{{{\color{numb}3}}}{1}
      {4}{{{\color{numb}4}}}{1}
      {5}{{{\color{numb}5}}}{1}
      {6}{{{\color{numb}6}}}{1}
      {7}{{{\color{numb}7}}}{1}
      {8}{{{\color{numb}8}}}{1}
      {9}{{{\color{numb}9}}}{1}
      {\{}{{{\color{delim}{\{}}}}{1}
      {\}}{{{\color{delim}{\}}}}}{1}
      {[}{{{\color{delim}{[}}}}{1}
      {]}{{{\color{delim}{]}}}}{1},
}

\usepackage{hyperref}

\title{Does Whisper Understand Swiss German? An Automatic, Qualitative and Human Evaluation}

\author{Eyal Liron Dolev \\
  Linguistics Center Zurich \\
  German Department \\
  University of Zurich \\
  \texttt{eyalliron.dolev@uzh.ch} \\\And
  Clemens Fidel Lutz \\
  Department of Computational Linguistics \\
  Phonetics \& Speech Sciences \\
  University of Zurich \\
  \texttt{clemensfidel.lutz@uzh.ch} \\\AND
  Noëmi Aepli \\
  Department of Computational Linguistics \\
  University of Zurich \\
  \texttt{noemi.aepli@uzh.ch}}

\begin{document}
\maketitle
\begin{abstract}
  Whisper is a state-of-the-art automatic speech recognition (ASR) model \cite{radford-2022-robust}. 
  Although Swiss German dialects are allegedly not part of Whisper's training data, preliminary experiments showed that Whisper can transcribe Swiss German quite well, with the output being a speech translation into Standard German.
  To gain a better understanding of Whisper's performance on Swiss German, we  systematically evaluate it using automatic, qualitative, and human evaluation. 
  We test its performance on three existing test sets: 
  SwissDial \citep{doganschönberger2021swissdial}, STT4SG-350 \citep{pluss-etal-2023-stt4sg}, and Swiss Parliaments Corpus \citep{plüss2021swiss}. In addition, we create a new test set for this work, based on short mock clinical interviews. 
  
  For automatic evaluation, we used word error rate (WER) and BLEU.
  In the qualitative analysis, we discuss Whisper's strengths and weaknesses and analyze some output examples. For the human evaluation, we conducted a survey with  28 participants who were asked to evaluate Whisper's performance.

  All of our evaluations suggest that Whisper is a viable ASR system for Swiss German, so long as the Standard German output is desired.
\end{abstract}

\section{Introduction}\label{sec:introduction}
Swiss German is the name of a group of Alemannic (High German) dialects spoken in German-speaking Switzerland by around 5.5 million people.\footnotemark{} 
\footnotetext{\href{https://www.bfs.admin.ch/asset/de/24205499}{Bundesamt für Statistik: Hauptsprachen seit 1910}, accessed on 23.04.2024}
German-speaking Switzerland displays a state of diglossia \citep{ferguson-1959-diglossia,rash1998german}, more specifically a medial diglossia: spoken contexts evoke Swiss German, written contexts evoke Standard German \citep{Kolde1983SprachkontakteIG,Haas+2004+81+110}.  
According to this principle, Swiss German is used as a spoken language in almost all settings, with the exception of some restricted, specific formal settings in which Standard German is spoken, e.g., on the news or at school, as well as a \emph{lingua franca} with non Swiss German speakers \citep{hogg-1984-diglossia}. 

Swiss German has no spoken standard variety and no written variety, and therefore no orthographic norms. In writing, Standard German is used. 
Thus, whenever spoken language (Swiss German) has to be written down, e.g.,   subtitles to a TV program or minutes of a meeting, Standard German is used. 
If Swiss German is written, it happens in situations that are conceptually spoken 
\citep[\enquote{\textit{konzeptionell mündlich;}}][]{koch-1994-schriftlichkeit} and which are situated on the immediacy-end of Koch and Oesterreichs's communication model 
\citep[\textit{Nähe-Distanz-Modell}, cf.][]{koch-1985-sprache}, e.g., ads or chat messages, cf. \citet{ueberwasser-2017-what}, which created a corpus of text messages written in Swiss German. 

To summarize, in German-speaking Switzerland there is a state of medial diglossia: the spoken language is Swiss German (a group of dialects with no standard variety); the written language is a Swiss vareity of Standard German. Swiss German and Standard German are, although genetically and systematically very close, two different languages, whereas only Standard German has a codified written form \citep{Berthele+2004+111+136}.
The task of putting down Swiss German speech to written form is, therefore, not a transcription task, but rather a translation task, translating Swiss German to Standard German.
This spoken--written juxtaposition of Swiss German and Standard German explains why almost all the automatic speech recognition efforts for Swiss German until now have dealt with Swiss German speech to Standard German text (see Section~\ref{sec:previous}).

Whisper is a state-of-the-art multilingual model for automatic speech recognition (ASR) \citep{radford-2022-robust}. Although Swiss German is not officially part of Whisper's training data\footnote{\url{https://github.com/openai/whisper}, accessed on 23.04.2024}, in preliminary trials, we observed that Whisper could recognize Swiss German quite well, with the output produced being Standard German. According to \citet{rudernewsletter24} most large language models (LLMs) have likely encountered some data for most languages available on the web, which is probably the case here too.


We intentionally refrain from attempting to fine-tune Whisper. 
Not only did \citet{sicard-etal-2023-spaiche}'s fine-tuning attempts of Whisper on Swiss German data worsen the model's performance; we find  Whisper's zero-shot performance on Swiss German, at this stage, already impressive and applicable. 
Before any costly GPU hours are spent in an attempt to improve Whisper, we think it should first be scrutinized and analyzed in its current state. 

In this work, we evaluate Whisper's performance on Swiss German audio in different settings and modes. 
We automatically evaluated Whisper on three large corpora, namely SPC \citep{plüss2021swiss}, STT4SG-350 \citep{pluss-etal-2023-stt4sg}, and SwissDial \citep{doganschönberger2021swissdial}, measuring word error rate (WER) and BLEU. 

To test Whisper on real-life spoken language, we created a new test set for which we translated into Standard German mock clinical interviews held in Swiss German. The total length of the interviews is approx.~30 minutes. 
To test Whisper's performance on this test set, we offer a qualitative analysis of Whisper's output and a human evaluation based on a survey (\(n=28\)).

\section{Previous Work}\label{sec:previous}
ASR for Swiss German is an ambiguous term. While the audio input to the system is always Swiss German, the text output can be: 
(a) dialectal writing -- loosely phonemic representation of Swiss German; 
(b) normalized writing -- transcriptions resembling standard German that are relatively consistent but distant from the acoustic signal \citep{nigmatulina-etal-2020-asr}; 
(c) Standard German translation.

In recent years, Swiss German has enjoyed a proper boom in the field of speech corpora, ASR and speech generation.
The first major corpus with Swiss German audio was ArchiMob, which includes dialectal as well as normalized writing \citep{samardzic-etal-2016-archimob,scherrer-2019-digitising}.
\citet{nigmatulina-etal-2020-asr} used the ArchiMob corpus to compare systems producing dialectal and normalized writing and concluded that performance is better with standardized writing.
\citet{doganschönberger2021swissdial} created SwissDial, a large corpus containing Standard German as well as Swiss German transcriptions in eight dialects.

Some work concentrated on ASR with Standard German speech translation and leveraged existing Transformer and XLS-R ASR models, fine-tuning them with Swiss German data.
\citet{plüss2021swiss} published the \enquote{Swiss Parliaments Corpus}, and experimented further with ASR models for Swiss German with Standard German output. 
\citet{pluss-etal-2022-sds} presented SDS-200, a corpus of Swiss German dialectal speech with Standard German text translations containing 200 hours of speech. 
They also experimented with training Transformer models and fine-tuning Wav2Vec2 XLS-R models on their data. 
Their best model (XLS-R) reached a WER of 21.6 and a BLEU score of 64.0.
Recently, \citet{pluss-etal-2023-stt4sg} presented the as-of-today largest corpus of Swiss German dialectal speech with Standard German text, containing 343 hours of speech. 
They fine-tuned a Wav2Vec2 XLS-R model on the corpus and reached a WER of 14.0 and a BLEU score of 74.7 on their test set.

Most recently, \citet{sicard-etal-2023-spaiche} turned to Whisper and tested it in a zero-shot setting on select Swiss German/Standard German test sets (SwissDial, SDS-200, SPC). 
Reportedly, their fine-tuning experiments on Whisper (medium version) worsened performance, leading the model to suffer from catastrophic forgetting.

\section{Test Sets}\label{sec:test-sets}
To evaluate Whisper's performance on Swiss German, we test it using WER and BLEU on three test sets: 
SwissDial \citep{doganschönberger2021swissdial}, Swiss Parliaments Corpus \citep{plüss2021swiss}, STT4SG-350 \citep{pluss-etal-2023-stt4sg}. We additionally created a new test set based on short Swiss German mock clinical interviews, which we additionally evaluate using a qualitative analysis and a human survey.

\subsection{Mock Clinical Interviews}\label{sec:mock-interviews}
This work serves as a preparation step towards a large longitudinal study in the field of suicide prevention.\footnote{\href{https://www.multicast.uzh.ch/en.html}{MULTICAST}}
During this study, patients from a Zurich-based psychiatric clinic will be interviewed several times. 
We test how reliable and viable Whisper is for transcribing/translating these interviews.

To this end, i.e., to test Whisper in a naturalistic and applied setting containing spontaneous speech, we used mock clinical interviews that were held in Swiss German and recorded for instructional and training purposes in a total length of approx.\,30 minutes. 
The interviews were recorded with three women interviewees using a lapel microphone\footnote{\href{https://rode.com/de/microphones/mobile/smartlav-plus}{RØDE smartLav+}} and simple convertible laptops.
We, the authors of this work, then translated these interviews into Standard German according to some basic translation guidelines we created to maintain consistency. 
We call this ad-hoc test set \enquote{Mock Clinical Interviews}.

This test set will be automatically evaluated using WER and BLEU as well as using a qualitative analysis to discuss Whisper's strengths and weaknesses in Swiss German, and a human evaluation, for which we conducted a survey (\(n=28\)).


\subsection{SwissDial}\label{sec:swissdial}
For the creation of SwissDial, eight speakers, speaking eight different dialects\footnotemark{} were asked to translate Standard German prompts to their own dialects and then record the translations. The prompts were made of sentences crawled from the internet, encompassing different text genres: news stories, Wikipedia articles, weather reports and short stories \citep{doganschönberger2021swissdial}. 
Because the prompts were translated into Swiss German by each of the speakers, sometimes greater departures from the Standard German source occur. 
See Figure~\ref{fig:ex-swissdial}, containing the first three dialect entries from the first entry in the corpus, for an example.

\begin{figure}
    \begin{lstlisting}[captionpos=b, label=lst:swissdial, language=json]
    {
      "id": 0,
      "de": "Derzeit ist er in \"Parasite\", dem Siegerfilm von Cannes, zu sehen.",
      "ch_sg": "Zur Ziit isch er in \"Parasite\", en Siegerfilm vo Cannes, zgseh.",
      "ch_be": "Momentan ischer in \"Parasite\" z gseh, em Siegerfium vo Cannes.",
      "ch_gr": "Derziit isch er in \"Parasite\", am Siegerfilm vu Cannes, z gseh.",
        ...
    }
    \end{lstlisting}
    \caption{The first three dialectal translations of the first entry in the SwissDial corpus.~The first word in the Standard German source (\enquote{\texttt{de}}), \emph{derzeit}, is translated differently in each dialect: \emph{zur ziit}, \emph{momentan}, \emph{derziit}.}
    \label{fig:ex-swissdial}
\end{figure}

As can be seen in Figure~\ref{fig:ex-swissdial}, the German word \emph{derzeit} \enquote{currently} was translated by the different dialect speakers as \emph{zur Ziit}, \emph{momentan} and \emph{derziit}, respectively. 
One cannot, however, expect that Whisper translates all of these different Swiss German words back to the Standard German original, especially considering that the Swiss German words each have a closer Standard German equivalent (\emph{zur Zeit}, \emph{momentan} and \emph{derzeit}, respectively).

To circumvent this problem and include prompts that are less likely to contain major departures from the source, which might unfairly fail Whisper when the produced output is compared to the original prompt, we created an ad-hoc test set: We calculated for each Standard German prompt and its respective dialectal translations the chrF score \citep{popovic-2015-chrf} using SacreBLEU's implementation \cite{post-2018-call}.
We then evaluated Whisper's performance on the 500 prompts with the best chrF scores for each dialect.

\footnotetext{The dialects of Zurich, Bern, Basel, Aargau, Grisons, St. Gallen, Lucerne, and the Walser}

\subsection{Swiss Parliaments Corpus}\label{sec:spc}
The Swiss Parliament Corpus (SPC) is a dataset containing sentences taken from speeches held at the Grand Council (Grosser Rat) of the Canton of Bern \citep{plüss2021swiss}.\footnote{The name of the corpus is thus a misnomer -- it is not a corpus representing the whole diversity of Swiss German.}
Almost all speakers hold their speeches in Bernese German.
For the creation of the corpus, \citet{plüss2021swiss} split the audio into segments, so-called sentences, whereas segments shorter than one second and longer than 15 seconds were discarded.
The corpus creators also made sure that the speech segments were unique within the set.
The speech segments were then force-aligned to the Standard German minutes (i.e., translations), which were created by the Canton of Bern.
The result is parliament speeches split into segments (sentences) with their corresponding Standard German translations from the minutes.
We tested Whisper on the test set part of the corpus\footnote{6 hours, 3332 segments}.

\subsection{STT4SG-350}
Like SPC, STT4SG-350\footnote{Standing for \enquote{Speech-to-text for Swiss German}} is a corpus containing single sentences of Swiss German speech with Standard German translations \citep{pluss-etal-2023-stt4sg}.
Unlike the former, STT4SG-350 includes an almost even split between seven different dialect regions.\footnotemark{}
The sentences produced by speakers were taken from Swiss newspapers and proceedings of two Swiss Parliaments.
Participants, who were recruited either via a crowdsourcing platform or academic or personal channels as well as news ads, self-reported their dialect region, age group, gender, and where they grew up and/or went to school.
The whole corpus consists of 343 hours of speech.
We tested Whisper on the test set part which contains 34 hours of speech in approx.~25k sentences.

\footnotetext{These seven regions are Basel, Bern, Grisons, Central Switzerland, East Switzerland, Valais and Zurich.}

\section{Evaluation}
\subsection{Automatic Evaluation}\label{subsec:automatic}
Usually, word error rate (WER) is used as a metric to automatically evaluate ASR systems.
However, the type of ASR for Swiss German that we evaluate in this work is Swiss German audio input with Standard German text output -- a speech translation task.
This means, as is generally the case in translation, that it is not uncommon for a sentence to have several possible translations. 
Standard German translations of Swiss German are, in that sense, no different, although in many cases, there are clear one-to-one correspondences in vocabulary and grammatical structures between Standard and Swiss German.
But when correspondences are ambiguous, the translator has to make a conscious decision on how to translate vocabulary or grammatical constructions.
For example, Swiss German only has one tense referring to past events -- the perfect. Standard German has, at least formally, two past tenses -- the perfect and the preterite.  The translator thus has to choose, according to context, how to translate the Swiss German perfect. 

This ambiguity in translation, a typical problem in evaluating machine translation systems, makes the usual metric used for ASR systems -- word error rate (WER) -- not unproblematic.
We thus additionally use BLEU \citep{papineni-etal-2002-bleu}, a typical metric used to evaluate machine translation systems.
This will also help compare the performance of Whisper to previous Swiss German ASR 
models, as previous work  also reports WER and BLEU. 


To compute WER, we used JiWER's\footnote{\url{https://github.com/jitsi/jiwer}, accessed on 23.04.2024} implementation. For BLEU we used SacreBLEU's implementation \citep{post-2018-call}.

\subsection{Qualitative \& Human Evaluation}
In addition to testing Whisper's performance on several datasets and evaluating  its performance automatically, 
we offer a qualitative and human evaluation of our Mock Clinical Interviews (see Section~\ref{sec:mock-interviews}). 
In the qualitative evaluation, we will show examples of Whisper's output, analyze errors, and 
shed some light on the strengths and weaknesses of Whisper's performance.

Our human evaluation, in which we recruited 28 people -- university students, colleagues, and acquaintances -- via personal channels to evaluate Whisper's output, offers more informative feedback about how humans perceive Whisper's output.



\section{Results: Automatic Evaluation}
\subsection{Mock Clinical Interviews}

We tested Whisper's large-v3 model on our test set (\enquote{Mock Clinical Intverviews}, see Section~\ref{sec:mock-interviews}). 
We compared Whisper's performance on continuous recordings versus short clips containing single speech segments. 
Given segmented clips, WER and BLEU scores were 0.37 and 44.19, respectively. 
With the continuous recordings, WER and BLEU scores were 0.33 and 52.03, respectively, see Table~\ref{tab:interviews}.
We conclude that Whisper performs better on longer, continuous recordings than on short clips.

\begin{table}
  \center
  \begin{tabular}{lcc}
  \toprule
  Mode & WER & BLEU \\
  \midrule
  Continuous recordings & \textbf{0.33} & \textbf{52.03} \\
  Segmented clips & 0.37 & 44.19 \\
  \bottomrule
  \end{tabular}
  \caption{Whisper's performance on our \emph{Mock Clinical Interviews} test set, comparing continous recordings vs.~segmented clips. Best results in bold.}
  \label{tab:interviews}
\end{table}

This comes, however, at a slight risk of hallucinations: Four out of sixteen transcriptions/translations generated by Whisper included one sentence that was not uttered in the original audio, see Section~\ref{sec:hallucinations} for more details.

\subsection{SPC, STT4SG \& SwissDial}\label{sec:spc-stt4sg-swissdial}
We further tested  Whisper's large-v3 model on the three other test sets:   SPC, STT4SG-350, and SwissDial (see Section~\ref{sec:test-sets}).
The results, compared to results reported by other works, can be seen in Table~\ref{tab:results-all}. 
We always picked the best result reported in each of the other works. 

\begin{table*}[!ht]
    \centering
    \begin{tabular}{lllccl}
    \toprule
        Test Set     & Model           & Mode & WER & BLEU &  \\ 
        \midrule
        Mock Interviews & Whisper large-v3 &      zero-shot & 0.372 & 44.3 & This work \\ 
        \midrule
        SPC     &Conformer & pre-trained  & 0.278 & 58.6 & \citet{plüss2021swiss} \\ 
        ~   &     \textbf{Wav2Vec2} & \textbf{fine-tuned} & \textbf{0.237} & \textbf{60.7} & \citet{schraner2022swiss}\\
        ~   &        Whisper large &              zero-shot & 0.332 & 55.6 & \citet{sicard-etal-2023-spaiche} \\
        ~     &        Whisper large-v3 &           zero-shot & 0.295 & 57.0 & This work \\ 
        \midrule
        
        STT4SG-350 & XLS-R & fine-tuned   &  0.153 & 72.2 & \citet{schraner2022swiss} \\
        ~          & \textbf{Wav2Vec2} & \textbf{fine-tuned}  & \textbf{0.140} & \textbf{74.7} & \citet{pluss-etal-2023-stt4sg} \\ 
                   & Whisper large-v3 & zero-shot & 0.230 & 63.1 & This work \\ 
        \midrule
        SwissDial & Whisper large & zero-shot & 0.294 & 56.2 & \citet{sicard-etal-2023-spaiche}\\
        ~ & \textbf{Whisper large-v3} & \textbf{zero-shot} & \textbf{0.230} & \textbf{61.0} & This work \\ 
        \bottomrule
    \end{tabular}
    \caption{WER (lower is better) and BLEU (higher is better) scores for our corpora, compared to results reported in previous works.}
    \label{tab:results-all}
\end{table*}

Whisper's latest large model, version 3, outperforms Whisper's previous model, as reported by \citet{sicard-etal-2023-spaiche}. 
However, fine-tuned Wav2Vec2 models on the SPC and the STT4SG-350 training sets outperform Whisper on the respective test sets, as reported by \citet{pluss-etal-2023-stt4sg} and \citet{schraner2022swiss}. 
Whisper does come close to the Conformer model pre-trained by \citet{plüss2021swiss} with a difference of only 1.7 \emph{p.p.}~and 1.6 \emph{p.p.}~in WER and BLEU, respectively.

For SPC, STT4SG-350, and SwissDial, we also computed WER and BLEU for each sentence separately and then computed the mean and standard deviation (so-called micro average). 
As can be seen in Table~\ref{tab:micro}, the standard deviations for WER and BLEU are quite big, ranging at 0.24--0.25 for WER and 27.95--32.24  for BLEU. This shows that Whisper's performance measured in WER and BLEU fluctuates considerably. For some sentences in STT4SG-350 for example, BLEU scores went up to 100. 
See also Figures~\ref{fig:wers} and~\ref{fig:bleus} in Appendix~\ref{app:plots}. 

\begin{table}[h!]
  \centering
  \begin{tabular}{lcc}
    \toprule
    Test Set & WER & BLEU \\
    \midrule
    SPC & 0.30  (0.24) & 54.01 (27.95) \\
    STT4SG-350 & 0.24 (0.25) & 60.61 (32.24) \\
    SwissDial & 0.25 (0.24) & 57.23 (31.51) \\
    \bottomrule
  \end{tabular}
  \caption{Mean and standard deviation WER and BLEU for the corpora when computed for each sentence separately.}
  \label{tab:micro}
\end{table}

It should be noted that the SPC corpus contains some considerable deviations between audio and reference translations, which were taken from the parliament's proceedings (see Section~\ref{sec:spc}). 
For instance, in one clip\footnote{\texttt{82495971-6523-4f96-be13-753b8bb564cf.flac}}, the heard audio is \emph{und das isch schlächt}. 
The reference translation is \enquote{Das ist schlecht}, excluding the coordinating conjunction \emph{und} \enquote{and}. 
Whisper perfectly transcribed this as \enquote{Und das ist schlecht}, but this is penalized with a WER score of 0.33.
It is not inconceivable, that the models trained by \citet{plüss2021swiss}  learned these deviating translations, which might explain their better performance on the SPC test set. 
As the case may be, comparing WER and BLEU scores for SPC between Whisper's performance and \citet{plüss2021swiss} may raise concerns, and its meaningfulness can and should be questioned.
For more examples of perfect output by Whisper penalized by diverging reference translations, see Table~\ref{tab:spc-examples} in Appendix~\ref{sec:appendix}.



For SwissDial, we also evaluated Whisper's performance on the different dialects.
As can be seen in Table~\ref{tab:swissdial}, the Grisons dialect has the best WER and BLEU scores; the Walser dialect has the worst scores.\footnote{The Walser dialect is also considered in Switzerland the most difficult to understand.}
Why Whisper performs differently on different dialects and which phonetic, phonological or grammatical traits  affect Whisper's performance should be more closely examined in future work.

\begin{table}[!ht]
  \centering
  \small
  \begin{tabular}{lcc}
  \toprule
      Dialect & WER & BLEU \\ \midrule
      Aargau & 0.272 & 55.40 \\ 
      Bern & 0.210 & 64.95 \\ 
      Basel & 0.209 & 63.24 \\ 
      Grisons & 0.169 & 69.99 \\ 
      Lucerne & 0.276 & 55.06 \\ 
      St.~Gallen & 0.209 & 64.03 \\ 
      Walser & 0.297 & 53.46 \\ 
      Zurich & 0.229 & 60.67 \\ \bottomrule
  \end{tabular}
  \caption{WER and BLEU scores for each dialect in the SwissDial corpus.}
  \label{tab:swissdial}
\end{table}

To conclude, we consider Whisper's results impressive, especially considering that it operates in a zero-shot setting. Its output is without doubt meaningful and useful.

\section{Qulatitative Analysis}\label{sec:qualitative}
\subsection{General Impression}
In general, we were genuinely impressed with Whisper's performance. 
The Standard German output corresponds in meaning and style to the Swiss German audio to almost the full extent. 
Whisper generated entire error-free passages that are fluent, consistent in style, retain the original word order and correspond fully to the original (see Table~\ref{tab:excerpts} in Appendix~\ref{app:excerpts} for examples).


However, some things are not always consistent. 
For example, the Swiss German perfect tense is translated sometimes as the Standard German perfect tense and sometimes as the preterite. 
The output switches inconsistently between the two past forms within the same passage. 
See Table~\ref{tab:perf-pret} in Appendix~\ref{app:excerpts} for examples.

We noticed that in certain cases, words are changed when translated to Standard German, even when the Swiss German word has an identical corresponding word in Standard German. 
One example of this is the Swiss German word \emph{lässig} which is translated to Standard German \emph{toll}, both meaning \enquote{cool, nice}. 
In this case, this is desired since in Standard German, \emph{lässig} means rather \enquote{casual, easy-going} -- Swiss German \emph{lässig} and Standard German \emph{lässig} are false friends. 
Another example is the translation of Swiss German \emph{Sache} to Standard German \emph{Dinge}, both meaning \enquote{things}, however, \emph{Dinge} is used mostly for tangible things and in the given contexts \emph{Sachen} would have been a better translation.

\subsection{Concise Style}
We notice that Whisper's translations are of a style that is more concise than the original. 
This is especially noticeable in the removal of modal particles and conjunctions:
Modal particles with little semantic content but with an information structural function like \emph{halt} or \emph{einfach} might disappear from the output. 
Conjunctions like \emph{dann} \enquote{then} or \emph{und} \enquote{and} are not always included. 
In one case, however, conjunctions and particles were hallucinated by Whisper.
Whisper deals then inconsistently with particles and conjunctions, mostly removing them but rarely also adding them by hallucination.

It is a known phenomenon that during translation, the explicitness of cohesive markers, such as the particles and conjunctions mentioned above, can shift \cite{blum1986shifts}.
Leaving out such markers, as evidenced in Whisper's output, can be seen as a case of implicitation, cf. \citet{lapshinova-koltunski-pollklaesener-przybyl:2022} (which refers to them as \enquote{discourse connectives}). If we assume that the target side of the training data was more concise and less explicit than the spoken Swiss German, then this would explain Whisper's behavior.

It should, however, be noted that such modal particles usually serve an information structural function \citep{musan2010Informationsstruktur}. Thus, they do not necessarily affect the truth value of an utterance and, therefore, have little influence on the overall meaning \citep{krifka2007BasicNotionsInformation}.
For examples of removed particles, see Table~\ref{tab:particles} in Appendix~\ref{app:excerpts}.

\subsection{Hallucinations}\label{sec:hallucinations}
Four out of sixteen transcriptions of whole conversations contained hallucinations -- a sentence that was generated by Whisper without a corresponding utterance in the source audio. 

In one conversation, in which the interviewee recounted the death of her mother, the following sentence was hallucinated: 

{\small
\begin{quote}
    \emph{Sie blieb nicht mehr in unserem pegen...
      Meine Frau, die ich so sehr liebte.} 
      (\enquote{She didn't remain in our GIBBERISH... 
      My wife, whom I loved so much.})
\end{quote}
}

In another conversation, a sentence was continued by a hallucination (marked in bold): 

{\small
\begin{quote}
\emph{Ähm ... Ja, jetzt bin ich immer noch etwas groggy, aber es geht etwas. \textbf{Ich bin ganz müde. Äh ... Okay, ich kann ... Äh ... Zuerst schon.}}, (\enquote{Ehm ... Yes, now I'm still somewhat groggy, but I'm managing. \textbf{I am really tired. 
  Eh ... 
  Okay, I can ...
  Eh ...
  Firstly.}})
    
\end{quote}
}

In a different case, a sentence was preceded by a hallucination (in bold): 

{\small 
\begin{quote}
    \emph{\textbf{Und ... äh ... Das hat mich sehr angestrengt. Äh ... }
Das hat mich sehr viel aufgewühlt.} 
(\enquote{\textbf{And ... 
eh ... 
That really strained me. 
Eh ...} That really upset me.})
\end{quote}
}

At the end of one conversation, \emph{Untertitel von S G}\footnote{Whisper's output included a real person's name, which we anonymize here for privacy reasons.} (\enquote{Subtitles by...}) was added.\footnotemark{}

\footnotetext{Obviously due to subtitles being part of the training data.}

We couldn't identify a pattern as to when and why hallucinations happen, 
but they seem to be a generally known problem with Whisper and are not specific to Swiss German audio.\footnotemark{} 
Therefore, users should be aware that there is a possibility of hallucinations being added and in doubt re-check the audio.

\footnotetext{A DuckDuckGo search for \enquote{openai whisper hallucination} returns many web pages discussing the issue.}

\section{Human Evaluation}\label{sec:human}
\subsection{Motivation}
Performance of ASR systems is usually reported in WER , cf.~\citet{radford-2022-robust,baevski2020wav2vec}. 
However, it is less meaningful for evaluating ASR for Swiss German with Standard German output since several outputs can be considered correct (see also Sections~\ref{sec:introduction} and~\ref{subsec:automatic}).
Therefore, BLEU established itself as a second metric reported in works on ASR for Swiss German \citep{pluss-etal-2022-sds,pluss-etal-2023-stt4sg,schraner2022swiss}.

BLEU is meaningful mostly as a relative metric, comparing several systems; as an absolute score, it is less meaningful. 
It has been the object of criticism since \citet{callison-burch-etal-2006-evaluating}. 
Even its significance as a relative metric use has been harshly criticized, with \citet{kocmi-etal-2021-ship} complaining that \enquote{the sole use of BLEU impeded the development of improved models leading to bad deployment decisions.}
If we acknowledge that language technology is made for human beings, then its most important evaluation should be what humans think about it. 
We therefore conducted a short survey to evaluate how human beings perceive Whisper's performance.

\subsection{Survey}
For the survey, we randomly picked three of the conversations recorded as Mock Clinical Interviews (see Section~\ref{sec:mock-interviews}) and extracted 119 sentence pairs consisting of our reference translation (sentence A) and Whisper's output (sentence B). 

In the evaluation task, participants were asked, on a scale of 1 to 5, to rate:

\begin{enumerate}[nosep]
  \item To what extent is the meaning of sentence A retained in sentence B?
  \item To what extent is sentence B fluent and natural?
\end{enumerate}

with 1 being the worst and 5 being the best score.
To assist the participants, each grade on the scale was given a verbal  description (see Table~\ref{tab:guidelines} in Appendix~\ref{app:guidelines} for details).
The participants were instructed to rate the fluency of sentence B (Whisper's output) independently from sentence A (reference) and to ignore punctuation marks.

Twenty-eight university students, colleagues, and acquaintances, who were recruited via personal channels, participated in the survey, all of them native speakers of German or Swiss German.
The mean scores for meaning and fluency among all raters were 4.358 $\pm$ 0.046 (\emph{SD} 0.239) and 4.39 $\pm$ 0.074 (\emph{SD} 0.387), respectively, out of a maximum of 5 points.
These scores suggest a very high human satisfaction with Whisper's performance.

\subsection{Worst Rated Sentences}
\begin{table*}[!ht]
  \centering
  \begin{tabular}{lp{0.4\linewidth}p{0.4\linewidth}c}
    \toprule
    ~ & Reference & Whisper & Mean \\
    \midrule
    1 & also meistens etwas \textbf{Kleines}, weil ich am Abend nicht so hunger habe & Also meistens etwas \textbf{Riesiges}, weil ich am Abend nicht so Hunger habe. & 2.31 \\
    2 & und habe meine Sachen genommen und dann \textbf{bin ich} auf den Bus gelaufen also \textbf{zur} Bushaltestelle & Ich nahm meine Sachen und \textbf{liess sie} auf den Bus. Also \textbf{an der} Bushaltestelle. & 2.50 \\
    3 & Auf das Tram gegangen, auf die \textbf{Forchbahn} und dann hierher gekommen. & Auf die \textbf{Furchtbahn} gingen. Dann kam ich hierher. & 2.54 \\
    4 & und dann bin ich bis zum \textbf{Stadelhofen} gefahren und habe ich noch die Tram genommen und dann bin ich hierher gekommen & Ich fuhr bis zum \textbf{Stadelhof} und nahm \textbf{ihn} aus der Rampe. Dann kam ich hierher. & 2.58 \\
    5 & Ja also ich bin um \textbf{viertel nach sechs} aufgestanden & Ich bin um \textbf{4.15} Uhr aufgestanden. & 2.65 \\
    6 & Da habe ich aber zuerst noch die \textbf{Schlummer-Taste} gedrückt zweimal. & 
    Ich habe zuerst die \textbf{Schlamasseltasche} gedrückt, zweimal. & 2.69 \\
    \bottomrule
  \end{tabular}
  \caption{The six sentences rated the worst in the human evaluation.}
  \label{tab:worst}
\end{table*}

In an attempt to uncover some of Whisper's weaknesses, we picked the six sentences with the lowest mean score across all raters, see Table~\ref{tab:worst}.

In sentence 1, the output includes the word \emph{Riesiges} \enquote{huge} instead of the original \emph{Kleines} \enquote{small}, which is the exact opposite. 
In sentence 2, the subject of the sentence changes from the original \emph{ich} \enquote{I} to \emph{sie} \enquote{they}, and the verb changes from \emph{genommen} \enquote{took} to \emph{liess} \enquote{let}, causing the output to diverge greatly in meaning from the reference. 
Also, the preposition changes from \emph{zur} in the reference to \emph{an der} in Whisper's output.
In sentence 3, the name of a train line in Zurich (\emph{Forchbahn}) is \enquote{misheard} as \emph{Furchtbahn} \enquote{fright train}.
Sentence 4 diverges greatly from the reference, with the use of the 3\textsuperscript{rd} person accusative pronoun \emph{ihn} without first introducing its referent, resulting in a genuine \emph{non sequitur}.
In sentence 5, the time mentioned in the original (\emph{viertel nach sechs} \enquote{quarter past six}) was changed to \emph{4.15}. 
Finally, in sentence 6, the word \emph{Schlummer-Taste} \enquote{snooze button} was misheard as \emph{Schlmasseltasche}, a gibberish word meaning \enquote{bad luck bag}.

There is no recurring pattern in these sentences. It seems, however, that the transcription of named entities (\emph{Forchbahn}, \emph{Stadelhofen}, cf.~sentences 3 and 4) and numbers (cf.~sentence 5) might result in errors.

\section{Conclusion}
We evaluated Whisper's performance on Swiss German audio using automatic evaluation (WER and BLEU), a qualitative analysis and a human survey. 
All three evaluation types are evidential of very high performance: 
WER and BLEU are on par or slightly below other systems (cf.~Table~\ref{tab:results-all}). 
The qualitative analysis revealed very high quality, retaining almost always the original meaning with only slight changes in style and some removal of cohesion markers such as particles and connectors. 
The human evaluation showed high human satisfaction (mean: 4.36/5.00, \emph{n}=28).

We are therefore convinced that Whisper can be used, as is and out-of-the-box, without any further adaptations, for transcribing Swiss German, providing that the desired output is Standard German and that some loss of cohesion markers is acceptable.

However, as with any AI-based tool, Whisper should be used with caution.
The qualitative analysis revealed some cases of changes in meaning, especially of numbers, as well as some hallucinations, though these were rare (one sentence in four out of sixteen 2-minute clips). 
In case of doubt, users should always refer to the original audio.
Nevertheless, for the task of transcribing large portions of Swiss German audio or as a first step in a pipeline with other tasks in mind, such as keyword extraction or sentiment analysis, we think Whisper is a helpful, useful, and viable ASR tool.


\section*{Limitations}
In this work, we evaluated Whisper's performance on Swiss German using automatic evaluation (WER and BLEU). 
We restricted ourselves to these metrics, since these are the metrics that are reported in previous works on ASR for Swiss German. 
Granted, other potentially better-suited metrics also come to mind, e.g., chrF \citep{popovic-2015-chrf} and BERTScore \citep{bert-score}. 
However, since models from previous works are not publicly available, we could not test them using different metrics besides WER and BLEU and had to rely on the scores reported in the respective works. 
Previous models not being publicly available also explains why we could not test the performance of previous models on our own test set (Mock Clinical Interviews), which would have been desirable.

\section*{Acknowledgements}
Eyal Liron Dolev is a doctoral student at the MULTICAST project, Swiss National Science Foundation project no.~205913. He would like to express his gratitude to his doctoral supervisor, Prof.~Guido Seiler, for allowing him the freedom to conduct this work.
This work was also supported by the Swiss National Science Foundation, project no.~191934.
We thank the FHNW Institute for Data Science for making the SPC corpus available, as well as SwissNLP for making the STT4SG-350 dataset available.
We also kindly thank the reviewers for their valuable feedback.

\bibliography{gsw-whisper-paper}

\appendix\section{Appendix}\label{sec:appendix}

\subsection{Examples of Whisper's Performance}\label{app:excerpts}
Table~\ref{tab:excerpts} offers two excerpts from Whisper's output for our \emph{Mock Clinical Interviews} test set (see Section~\ref{sec:mock-interviews}). The excerpts exemplify Whisper's consistent high-quality performance over a longer passage of spontaneous and continuous speech.

\begin{table*}
  \begin{tabular}{p{0.45\linewidth}p{0.45\linewidth}}
    \toprule
    Reference translation & Whisper \\
    \midrule
    und es ist einfach so mega heimelig irgendwie also so & Es ist einfach so mega heimelig irgendwie. \\
    es gibt mir so ein warmes Gefühl & Es gibt mir so ein warmes Gefühl. \\
    und mir gefällt auch wie sie sich anziehen dort also, & Und mir gefällt auch, wie sie sich dort anziehen.  \\
    also ich finde es auch so schön es zu sehen, & Ich finde es auch so schön, es zu sehen.\\
    nicht nur weil es lustig ist und
    beruhigend sondern auch, weil es schön ist zum anschauen &
    Nicht nur, weil es lustig und beruhigend ist, sondern auch, weil es schön ist, es anzuschauen. \\
    \midrule
    also man hat Einzeltherapie zum Beispiel, oder Gruppentherapie mit anderen zusammen & also man hat Einzeltherapie zum Beispiel oder Gruppentherapie mit anderen zusammen \\
    oder Musiktherapie, Maltherapie, oder auch so Entspannungsgruppen & oder Musiktherapie, Maltherapie oder auch so Entspannungsgruppen. \\
    teilweise kann man auch selber Sport machen wenn man das will, also nicht in der Gruppe sondern alleine & Teilweise kann man auch selber Sport machen, wenn man das will, also nicht in der Gruppe, sondern alleine, \\
    oder irgendwie so Walkinggruppen, oder so Achtsamkeitsgruppen, wo man in die Natur geht.  & oder irgendwie so Walking-Gruppen oder so Achtsamkeitsgruppen, wo man in die Natur geht. \\
  \bottomrule    
  \end{tabular}
  \caption{Excerpts of Whisper's performance on continuous speech from our \enquote{Mock Clinical Interviews} test set (the segmentation into sentences is only for the sake of readability). These excerpts are evidential of Whisper high-quality performance.}
  \label{tab:excerpts}
\end{table*}

Table~\ref{tab:perf-pret} offers a speech excerpt from an interview in which the interviewee describes a past narrative (morning routine). 
It is an example of how Whisper inconsistently translates the Swiss German perfect sometimes as the Standard German preterite and sometimes as the Standard German perfect, cf.~Section~\ref{sec:qualitative}.

\begin{table*}
  \centering
  \begin{tabular}{p{0.45\linewidth}p{0.45\linewidth}}
  \toprule
  Swiss German & Whisper \\
  \midrule
   Denn \textbf{bin} i richtig \textbf{ufgstandfe} & Dann \textbf{bin} ich richtig \textbf{aufgestanden}. \\
  Dänn \textbf{bin} i go dusche \textbf{ggange} & Ich \textbf{ging} duschen.\\
  Dänn \textbf{han} i mi \textbf{aazoge} & Dann \textbf{zog} ich mich an. \\
 Dänn mine chatz no fuetter \textbf{gää}, will si di ganzi ziit am maue \textbf{gsi} \textbf{isch} und unbedingt \textbf{het} \textbf{welle} esse & Ich \textbf{gab} meinen Katzen Futter, weil sie die ganze Zeit am Mauen \textbf{waren} und essen \textbf{wollten}. \\
 Dänn \textbf{bin} i mit ire id stube abe ggange & Dann \textbf{ging} ich mit ihr in den Wohnzimmer.\\
 Dänn \textbf{han} \emph{ich} öppis \textbf{ggässe} dänn \textbf{het} \emph{sii} öppis \textbf{ggässe} & Ich \textbf{habe} etwas \textbf{gegessen}, dann \textbf{hat} sie etwas \textbf{gegessen}.\\
 Und dänn \textbf{bin} i wider uffe go zäh putze & Ich \textbf{ging} wieder hoch, um die Zähne zu putzen.\\
\bottomrule
  \end{tabular}
  \caption{Excerpt from a conversation in which the interviewee recounted what they did the same morning.
  It exemplifies how Whisper switches between perfect and preterite in Standard German. 
  The input is always in the perfect tense. 
  Perfect/preterites are marked in bold.}
  \label{tab:perf-pret}
\end{table*}

\begin{table*}
  \centering
  \begin{tabular}{p{0.45\linewidth}p{0.45\linewidth}}
    \toprule
    Reference & Hypothesis \\
    \midrule
    \enquote{weil ich \textbf{dann} \textbf{halt} wieder auf mich gestellt bin.} &
    \enquote{weil ich wieder auf mich gestellt bin.} \\
    \enquote{und darum ist es \textbf{ein bisschen} beides.} & \enquote{Darum ist es beides.} \\
    \enquote{\textbf{ja und} ich find's \textbf{einfach nur} spannend}" & \enquote{Ich finde es spannend} \\
    \enquote{\textbf{Halt irgend so} eine Einschlafmeditation von einer Person...} & \enquote{Eine Einschlafmeditation von einer Person...} \\
    \enquote{und er bekommt 50'000 Franken} & \enquote{und \textbf{dann} bekommt man \textbf{irgendwie noch} 50'000 Franken}" \\
    \bottomrule
  \end{tabular}
  \caption{Examples for the removal of particles and conjunctions in Whisper's output. 
  Words in bold are particles/conjunctions missing in the reference/hypothesis.}
  \label{tab:particles}
\end{table*}

\subsection{Human Evaluation Guidelines}\label{app:guidelines}
Table~\ref{tab:guidelines} shows the rating guidelines for the raters in the human evaluation survey, cf.~Section~\ref{sec:human}.
\begin{table*}
  \centering
  \begin{tabular}{cl}
  \toprule
  ~& \textbf{Sinn -- Ist der originale Sinn beibehalten? Entspricht Satz B Satz A?} \\
  5 & Entspricht sinngemäss voll und ganz dem Original \\
  4 & Etwas ist verloren gegangen, die Bedeutung ist aber im grossen und ganzen gleich \\
  3 & Stimmt teilweise, aber nicht in allen Teilen \\
  2 & Entspricht kaum noch dem originalen Sinn \\
  1 & Gar nicht \\
\midrule
~ & \textbf{Flüssigkeit. Bezogen auf Satz B – ist das gutes Deutsch?} \\
 5 & Ja, voll und ganz. Natürlich und einwandfrei. \\
4 & Relativ flüssig \\
3 & Nicht ganz flüssig, etwas merkwürdig \\
2 & Kaum akzeptabel \\
1 & Inakzeptabel \\
\bottomrule
  \end{tabular}
  \caption{Rating guidelines for the raters participating in the survey of human evaluation.}
  \label{tab:guidelines}
\end{table*}

\subsection{SPC Examples}
Table~\ref{tab:spc-examples} shows some examples of diverging reference translations that unjustly penalize Whisper's performance, cf.~Section~\ref{sec:spc-stt4sg-swissdial}.

\begin{table*}
  \centering
  \small
  \begin{tabular}{p{0.28\linewidth}p{0.28\linewidth}p{0.28\linewidth}c}
    \toprule
     Swiss German Audio & SPC Reference & Whisper & WER \\
     \midrule
     ...nachhinei muss me döt de iibürgerigswillige sägge, er het scho... & 
     So muss den Einbürgerungswilligen im Nachhinein gesagt werden: & Nachhinein muss man den Einbürgerungswilligen sagen, er hat schon & 0.75 \\
     Dir wüssed scho vo de römerziite her & Aus Römerzeiten wissen Sie schon: &	Ihr wisst schon von den Römerzeiten her, & 1.4 \\
     ...und das isch schlächt & Das ist schlecht. &	Und das ist schlecht.	& 0.33 \\
     Während acht jahr isch s in betriib gsi & Während acht Jahren wurde es betrieben. &	Während acht Jahren war es in Betrieb. &	0.5 \\
     ...u es het halt i Gotts name oo mitem finanzielle z tüe... & Und es hat halt auch wirklich mit dem finanziellen Aspekt zu tun. & Und es hat halt in Gottes Namen auch mit dem Finanziellen zu tun. & 0.42 \\
     Töu vo euch erinnere sech müglicherwiis aa experiment ir physik oder chemie & Manche von Ihnen erinnern sich möglicherweise an missglückte Experimente in Physik oder Chemie. & Ein Teil von euch erinnert sich möglicherweise an Experimente in Physik oder Chemie. & 0.38 \\
    \bottomrule
  \end{tabular}
  \caption{Examples for perfect performance of Whisper penalized by strongly divergent reference translations in the SPC corpus.}
  \label{tab:spc-examples}
\end{table*}

\subsection{Plots}\label{app:plots}
\begin{figure*}
  \centering
  \includegraphics[width=0.9\linewidth]{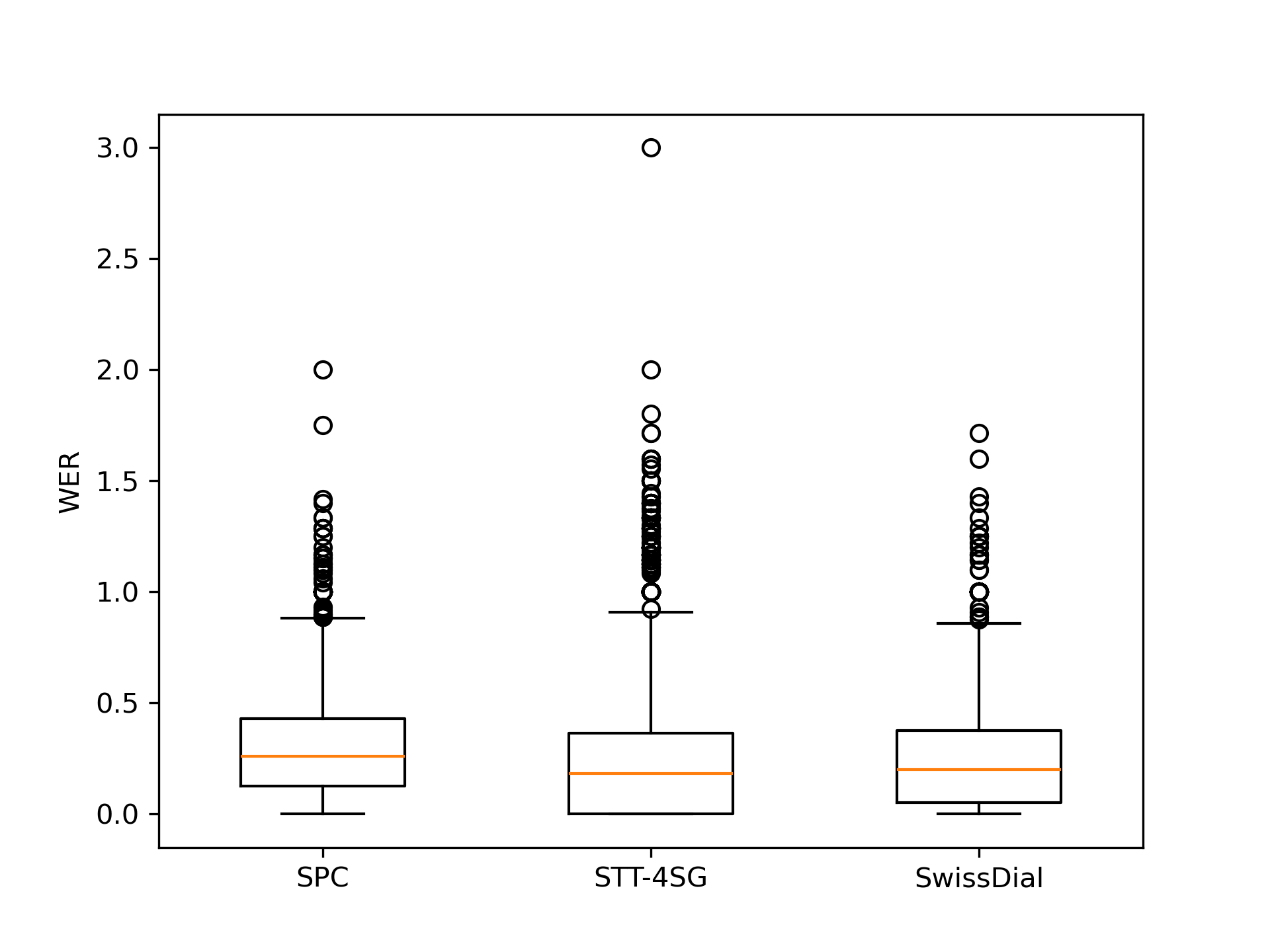}
  \caption{Distribution of WER scores for each corpus.}
  \label{fig:wers}
\end{figure*}

\begin{figure*}
  \centering
  \includegraphics[width=0.9\linewidth]{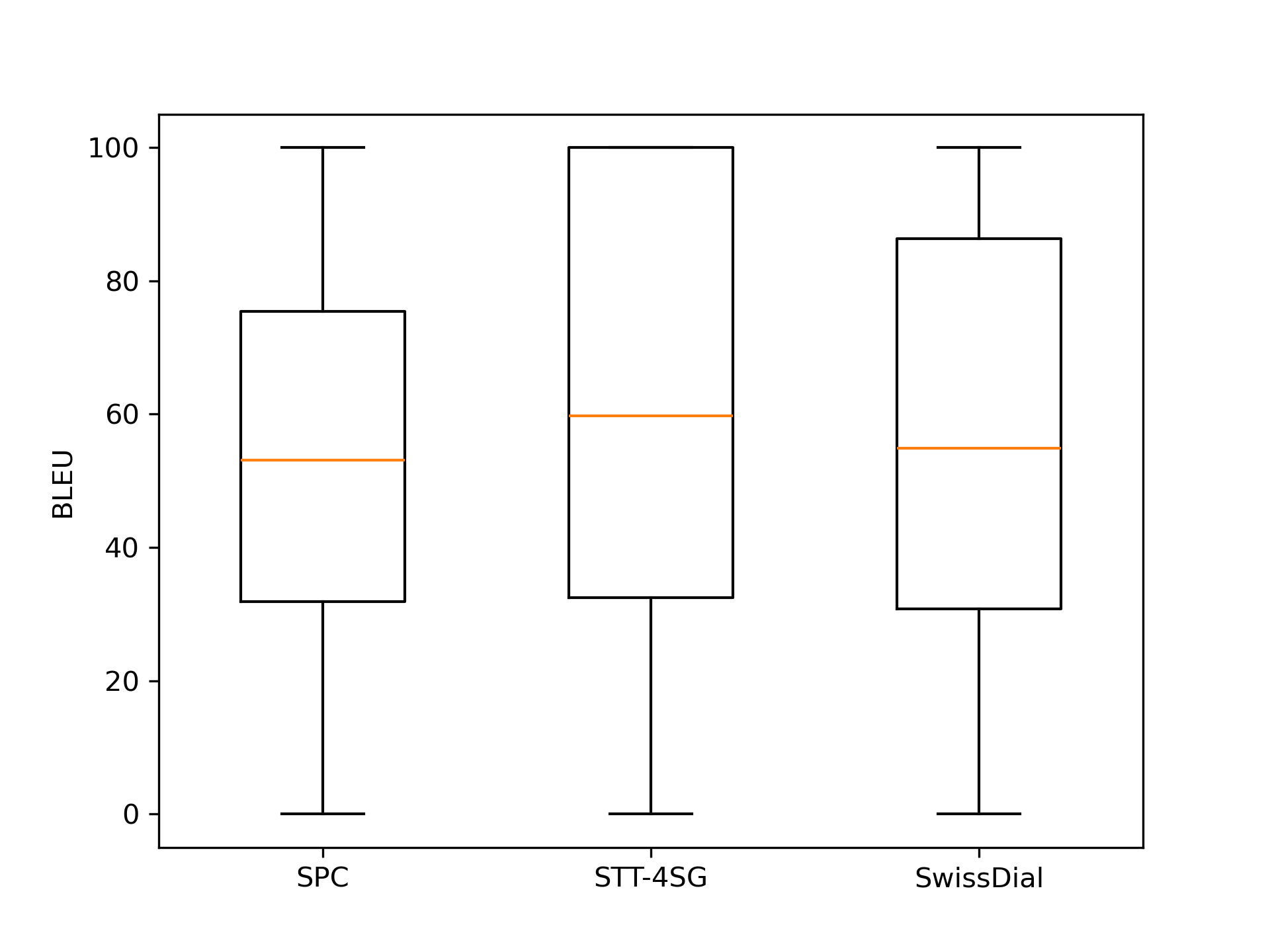}
  \caption{Distribution of BLEU scores for each corpus.}
  \label{fig:bleus}
\end{figure*}

Figures~\ref{fig:wers} and~\ref{fig:bleus} contain boxplots of the distribution of WER and BLEU scores of Whisper's performance on the three test sets: SPC, STT4SG and SwissDial. They show that Whisper's performance measured in WER and BLEU fluctuates considerably; for some sentences in STT4SG-350 for example, BLEU scores went up to 100. 

\end{document}